%% file: main.tex
\pdfoutput=1

\documentclass[11pt]{article}

\usepackage[table]{xcolor}

\usepackage[preprint]{acl}
\usepackage{graphicx}
\usepackage{times}
\usepackage{latexsym}

\usepackage[T1]{fontenc}

\usepackage[utf8]{inputenc}

\usepackage{microtype}
\usepackage{pdfpages}
\usepackage{inconsolata}
\usepackage{amssymb}
\usepackage{pifont}
\usepackage{multirow}
\usepackage{array}
\usepackage{float}
\usepackage{pdfpages}
\usepackage{algorithm}
\usepackage{algpseudocode}

\def\codebaselink{http://www.overleaf.com}

\newcommand{\tool}{\texttt{MapCoder}~}
\newcommand{\toolnospace}{\texttt{MapCoder}}

\def\ourapproach{\toolnospace}

\title{\toolnospace: {Multi-Agent Code Generation} \\ {for Competitive Problem Solving}}

\author{
Md. Ashraful Islam$^1$, \ Mohammed Eunus Ali$^1$, \ Md Rizwan Parvez$^2$ \\
$^1$Bangladesh University of Engineering and Technology \\
$^2$Qatar Computing Research Institute (QCRI) \\
\{mdashrafulpramanic, mohammed.eunus.ali\}@gmail.com, mparvez@hbku.edu.qa 
}

\begin{document}
\maketitle

\input{sections/abstract}
\input{sections/intro}
\input{sections/related-work}
\input{sections/our-approach}
\input{sections/experimental-setup}

\input{sections/results}

\input{sections/ablation-study}
\input{sections/conclusion}
\input{sections/limitations}

\input{sections/ack}

\bibliography{custom}

\input{appendix/appendix}

\end{document}

%% file: sections/abstract.tex
\begin{abstract}
Code synthesis, which requires a deep understanding of complex natural language (NL) problem descriptions, generation of code instructions for complex algorithms and data structures, and the successful execution of comprehensive unit tests, presents a significant challenge. Thus, while large language models (LLMs) demonstrate impressive proficiency in natural language processing (NLP), their performance in code generation tasks remains limited. In this paper, we introduce a new approach to code generation tasks leveraging the multi-agent prompting that uniquely replicates the full cycle of program synthesis as observed in human developers. Our framework, {\bf \toolnospace}, consists of four LLM agents specifically designed to emulate the stages of this cycle: recalling relevant examples, planning, code generation, and debugging. After conducting thorough experiments, with multiple LLMs ablations and analyses across eight challenging competitive problem-solving and program synthesis benchmarks—\tool showcases remarkable code generation capabilities, achieving their new state-of-the-art (pass@1) results—({HumanEval \bf 93.9\%}, {MBPP \bf 83.1\%}, {APPS \bf 22.0\%}, {CodeContests \bf 28.5\%}, and {xCodeEval \bf 45.3\%}).
Moreover, our method consistently delivers superior performance across various programming languages and varying problem difficulties. We open-source our framework at \url{https://github.com/Md-Ashraful-Pramanik/MapCoder}.




\end{abstract}

%% file: sections/intro.tex
\section{Introduction}
\label{sec:intro}
Computer Programming has emerged as an ubiquitous problem-solving tool that brings tremendous benefits to every aspects of our life \cite{li2022competition, parvez-etal-2018-building, knuth1992literate}. To maximize programmers’ productivity, and enhance accessibility, automation in program synthesis is paramount. With the growth of LLMs, significant advancements have been made in program synthesis—driving us in an era where we can generate fully executable code, requiring no human intervention~\cite{chowdhery2022palm, nijkamp2022codegen}.

Despite LLMs' initial success and the scaling up of model size and data, many of these models  still struggle to perform well on complex problem-solving tasks, especially in competitive programming problems \cite{austin2021program}. To mitigate this gap, in this paper, we introduce {\bf \toolnospace}: a {\bf M}ulti-{\bf A}gent {\bf P}rompting Based Code Generation approach that can seamlessly synthesize solutions for competition-level programming problems. 

{Competitive programming} or {competition-level code generation}, often regarded as the pinnacle of problem-solving, is an challenging task. It requires a deep comprehension of NL problem descriptions, multi-step complex reasoning beyond mere memorization, excellence in  algorithms and data structures, and the capability to generate substantial code that produces desired outputs aligned with comprehensive test cases \cite{khan2023xcodeeval}. 

\begin{figure*}
    \centering
    \vspace{-4mm}
    \includegraphics[width=0.99\textwidth]{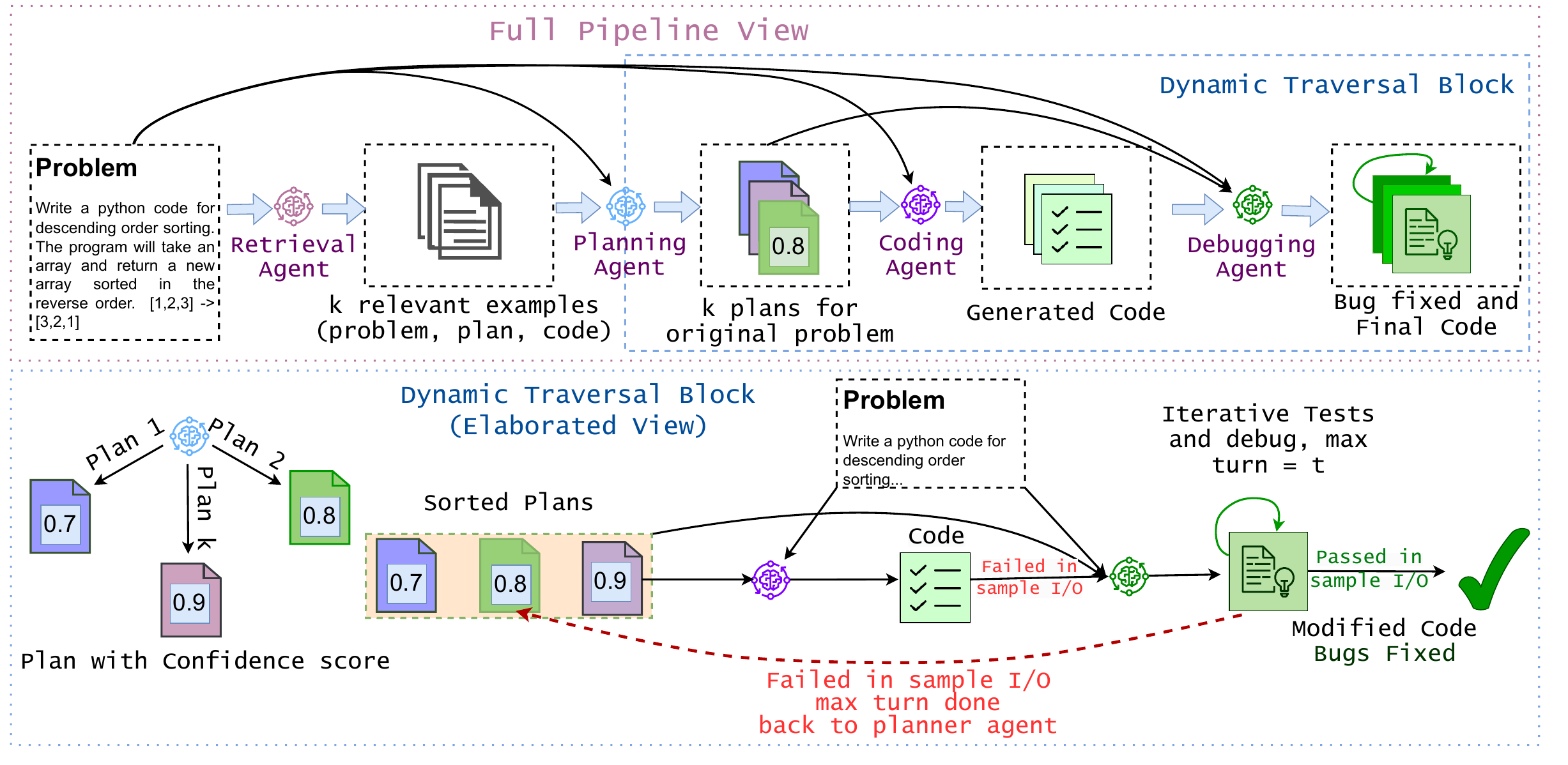}
    \vspace{-4mm}
    \caption{Overview of \tool (top). It starts with a retrieval agent that generates relevant examples itself, followed by planning, coding, and iterative debugging agents. Our dynamic traversal (bottom) considers the confidence of the generated plans as their reward scores and leverages them to guide the code generation accordingly.}
    \label{fig:map-coder-overview}
    \vspace{-6mm}
\end{figure*}  

Early approaches utilizing LLMs for code generation employ a direct prompting approach, where LLMs generate code directly from problem descriptions and sample I/O \cite{chen2021codex}. Recent methods like chain-of-thought \citep{CoT} advocates modular or pseudo code-based generation to enhance planning and reduce errors, while retrieval-based approaches such as  \citet{parvez2021retrieval} leverage relevant problems and solutions to guide LLMs' code generations. However, gains in such approaches remains limited in such a complex task like code generation where LLMs' generated code often fails to pass the test cases and they do not feature bug-fixing schema \cite{ridnik2024code}. 

A promising solution to the above challenge is self-reflection~\cite{shinn2023reflexion, chen2022codet}, which iteratively evaluates the generated code against test cases, reflects on mistakes and modifies accordingly. However, such approaches have limitations too. Firstly, while previous studies indicate that superior problem-solving capabilities are attained when using in-context exemplars \citep{shum-etal-2023-automatic, zhang2022automatic, CoT} or plans \citep{jiang2023self}, these approaches, during both code generation and debugging, only leverage the problem description itself in a zero-shot manner. Consequently, their gains can be limited.

To confront the above challenge, we develop \tool augmenting the generation procedure with possible auxiliary supervision. We draw inspiration from human programmers, and how they use various signals/feedback while programming. The human problem-solving cycle involves recalling past solutions, planning, code writing, and debugging. \tool imitates these steps using LLM agents - retrieval, planning, coding, and debugging. In contrast to relying on human annotated examples, or external code retrieval models,  we empower our retrieval agent to autonomously retrieve relevant problems itself \cite{yasunaga2023large}.
Moreover, we design a novel structured pipeline schema that intelligently cascades the LLM agents and incorporates a dynamic iteration protocol to enhance the generation procedure at every step. Figure \ref{fig:map-coder-overview} shows an overview of our approach, \tool.

Additionally, existing iterative self-reflection methods rely on extra test cases generated by LLM agents (e.g., AgentCoder \cite{huang2023agentcoder}, LATS \cite{zhou2023languageLATS}, self-reflection \cite{shinn2023reflexion}) or external tools, compounding the challenges. Test case generation is equally challenging as code generation \cite{pacheco2007feedback}, 
and incorrect test cases can lead to erroneous code. Blindly editing code based on these test cases can   undermine problem-solving capabilities.  For instance, while self-reflection boosts GPT-4's performance on the HumanEval dataset, it drops by 3\% on the MBPP dataset \cite{shinn2023reflexion}. Upon identification, to validate this, on the HumanEval dataset itself, we replace their GPT-4 with ChatGPT, and see that model performance drops by 26.3\%. Therefore, our debugging agent performs unit tests and bug fixing using only the sample I/O, without any artifact-more plausible for real-world widespread adoption. 

We evaluate \tool on seven popular programming synthesis benchmarks including both basic programming like HumanEval, MBPP and challenging competitive program-solving benchmarks like APPS, CodeContests and xCodeEval. With multiple different LLMs including ChatGPT, GPT-4, and Gemini Pro, our approach significantly enhances their problem-solving capabilities - consistently achieving new SOTA performances, outperforming strong baselines like Reflexion \citep{shinn2023reflexion}, and AlphaCodium \citep{ridnik2024code}. Moreover, our method consistently delivers superior performance across various programming languages and varying problem difficulties. Furthermore, with detailed ablation studies, we analyze \tool to provide more insights. 

%% file: sections/related-work.tex
\section{Related Work}

\textbf{Program Synthesis:}
Program synthesis has a long standing history in AI systems \citep{Zohar71}. A large number of prior research attempted to address it via search/data flow approaches \cite{li2022competition, parisotto2017neural, polozov2015flashmeta, gulwani2011automating}. 
LMs, prior to LLMs, attempt to generate code by fine-tuning (i.e., training) neural language models \citep{wang2021codet5, ahmad2021unified, feng2020codebert, parvez-etal-2018-building,cmu_code_gen,deep_net_for_source_code,code_gen_parsing, naturalnessofsoft}, conversational intents or data flow features \citep{andreas-etal-2020-task, yu-etal-2019-cosql}. \smallskip\\
\noindent{\bf Large Language Models:}
Various  LLMs have been developed for Code synthesis~\citep{alphacode, fried2022incoder, chen2021evaluating, austin2021program, nijkamp2022codegen, allal2023santacoder}. Recent open source LLMs include Llama-2 \citep{touvron2023llama}, CodeLlama-2 \citep{roziere2023code}, Mistral \citep{jiang2023mistral} Deepseek Coder \cite{guo2024deepseek}, MoTCoder \cite{li2023motcoder} that are capable of solving many basic programming tasks. 

\input{tables/feature-comparison}

\noindent{\bf Prompting LLMs:}
As indicated in Section \ref{sec:intro},  LLM prompting can be summarized into three categories: retrieval \cite{yasunaga2023large, parvez-etal-2023-retrieval, parvez2021retrieval}; planning \citep{wei2022chain, jiang2023self}; debugging \citep{ridnik2024code, chen2023teaching, chen2022codet, le2022coderl} apart from the direct code generation approaches. In contrast, we combine all these paradigms and bridge their gaps (See Table \ref{tab:feature-compare-table}). Among others, in different contexts of generic problem-solving,  Tree-of-thoughts \cite{yao2023tree}, and Cumulative reasoning \cite{zhang2023cumulative} approaches consider a tree traversal approach to explore different sub-steps towards a solution while our code generation approach mirrors the human programming cycle through various LLM agents. Notably, our traversal does not rely on sub-steps toward the solution but instead utilizes different forms of complete solutions.

%% file: tables/feature-comparison.tex

\begin{table}[h]
    \centering
    \begin{tabular}{c}
    \hspace*{-0.35cm}
    \includegraphics[width=0.49\textwidth]{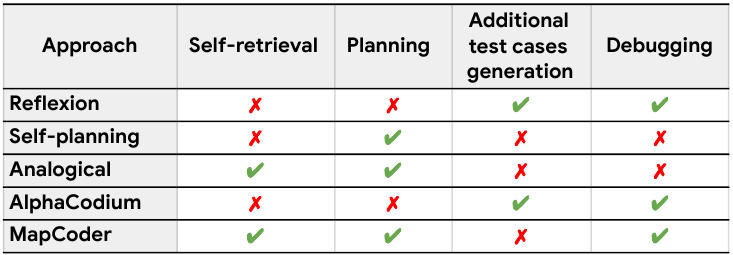}
    \end{tabular}
    \vspace{-3mm}
    \caption{Features in code generation prompt techniques.}
    \label{tab:feature-compare-table}
    \vspace{-2mm}
\end{table}

%% file: sections/our-approach.tex
\section{\tool}
\label{sec:mapcoder}
Our goal is to develop a multi-agent code generation approach for competitive problem-solving. In order to do so, our framework, \toolnospace, replicates the human programming cycle through four LLM agents - retrieval, plan, code, and debug. We devise a pipeline sequence for \toolnospace, intelligently cascading the agents in a structured way and enhancing each agent's capability by augmenting in-context learning signals from previous agents in the pipeline. However, not all the agent responses/outputs are equally useful. Therefore, additionally, \tool features an adaptive agent traversal schema to interact among corresponding agents dynamically, iteratively enhancing the generated code by, for example, fixing bugs, while maximizing the usage of the LLM agents.  In this section, we first discuss the agents (as per the pipeline), their prompts, and interactions, followed by the dynamic agent traversal protocol in \tool towards code generation for competitive problem-solving. 

\subsection{Retrieval Agent}
\label{subsec:our-algo-agent-1}
Our first agent, the \emph{Retrieval Agent}, recalls past relevant problem-solving instances, akin to human memory. It finds $k$ (user-defined) similar problems without manual crafting or external retrieval models. Instead, we leverage the LLM agent itself, instructing it to generate such problems. Our prompt extends the analogical prompting principles \cite{yasunaga2023large}, generating examples and their solutions simultaneously, along with additional metadata (e.g., problem description, code, and plan) to provide the following agents as auxiliary data. We adopt a specific sequence of instructions, which is crucial for the prompt's effectiveness. In particular, initially, we instruct the LLM to produce similar and distinct problems and their solutions, facilitating problem planning reverse-engineering. Then, we prompt the LLM to generate solution code step-by-step, allowing post-processing to form the corresponding plan. Finally, we direct the LLM to generate relevant algorithms and provide instructional tutorials, enabling the agent to reflect on underlying algorithms and generate algorithmically similar examples.

\subsection{Planning Agent}
\label{subsec:our-algo-agent-2}

The second agent, the \emph{Planning Agent}, aims to create a step-by-step plan for the original problem. Our \emph{Planning Agent} uses examples and their plans obtained from the retrieval agent to generate plans for the original problem. A straightforward approach would be to utilize all examples collectively to generate a single target plan. However, not all retrieved examples hold equal utility. Concatenating examples in a random order may compromise the LLM's ability to generate accurate planning. For instance, \citet{xu2023rereading} demonstrated that even repeating more relevant information (e.g., query) towards the end of the in-context input aids LLM reasoning more effectively than including relatively less relevant contexts. A similar conclusion of "separating noisy in-context data" can also be drawn from the state-of-the-art retrieval augmented generation approaches like \citet{filco}. Therefore, we generate a distinct target plan for each retrieved example. Additionally, multiple plans offer diverse pathways to success. 

\begin{figure}
    \centering
    \hspace*{-0.3cm}
    \includegraphics[width=0.49\textwidth]{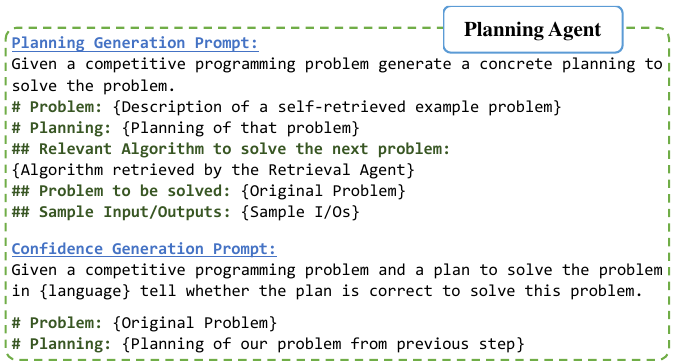}
    \vspace{-6mm}
    \caption{Prompt for \emph{Planning Agent}.} 
    \label{fig:prompt-agent-2-short}
    \vspace{-4mm}
\end{figure}

To help the generation steps in the following agents with the utility information for each plan, our designed prompt for the planning agent asks the LLM to generate both plans and a confidence score. Figure \ref{fig:prompt-agent-2-short} shows our prompt got this agent.

\smallskip
\subsection{Coding Agent}
\label{subsec:our-algo-agent-3}
Next is the \emph{Coding Agent}. It takes the problem description, and a plan from the \emph{Planning Agent} as input and translates the corresponding planning into code to solve the problem. During the traversing of agents, \emph{Coding Agent} takes the original problem and one particular plan from the \emph{Planning Agent}, generates the code, and test on sample I/O. If the initial code fails, the agent transfers it to the next agent for debugging. Otherwise, predicts that as the final solution.

\smallskip
\subsection{Debugging Agent}
\label{subsec:our-algo-agent-4}
\begin{figure}[h]
    \centering
    \hspace*{-0.3cm}
    \includegraphics[width=0.49\textwidth]{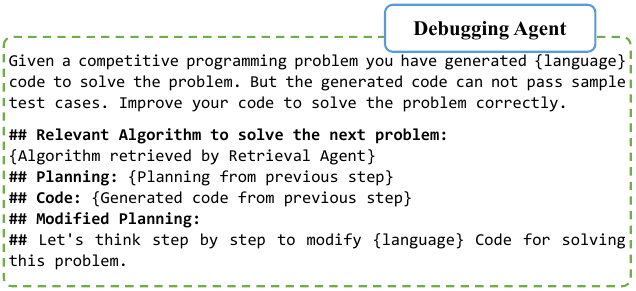}
    \vspace{-6mm}
    \caption{Prompt for \emph{Debugging Agent}.}
    \label{fig:prompt-agent-4-short}
    \vspace{-4mm}
\end{figure}

Finally, the \emph{Debugging Agent} utilizes sample I/O from the problem description to rectify bugs in the generated code. Similar to humans cross-checking their plan while fixing bugs, our pipeline supplements the \emph{Debugging Agent} with plans from the \emph{Planning Agent}. This plan-derived debugging significantly enhances bug fixing in \toolnospace, underscoring the pivotal roles played by both the \emph{Debugging Agent} and the \emph{Planning Agent} in the generation process. We verify this in Section \ref{sec:ablation-study}. For each plan, this process is repeated $t$ times. The prompt for this step is illustrated in Figure \ref{fig:prompt-agent-4-short}. Note that, different from Reflexion \cite{shinn2023reflexion} and AlphaCodium \cite{ridnik2024code}, our \emph{Debugging Agent}  does not require any additional test case generation in the pipeline. 

\begin{figure*}[h]
    \centering
    \includegraphics[width=0.99\textwidth]{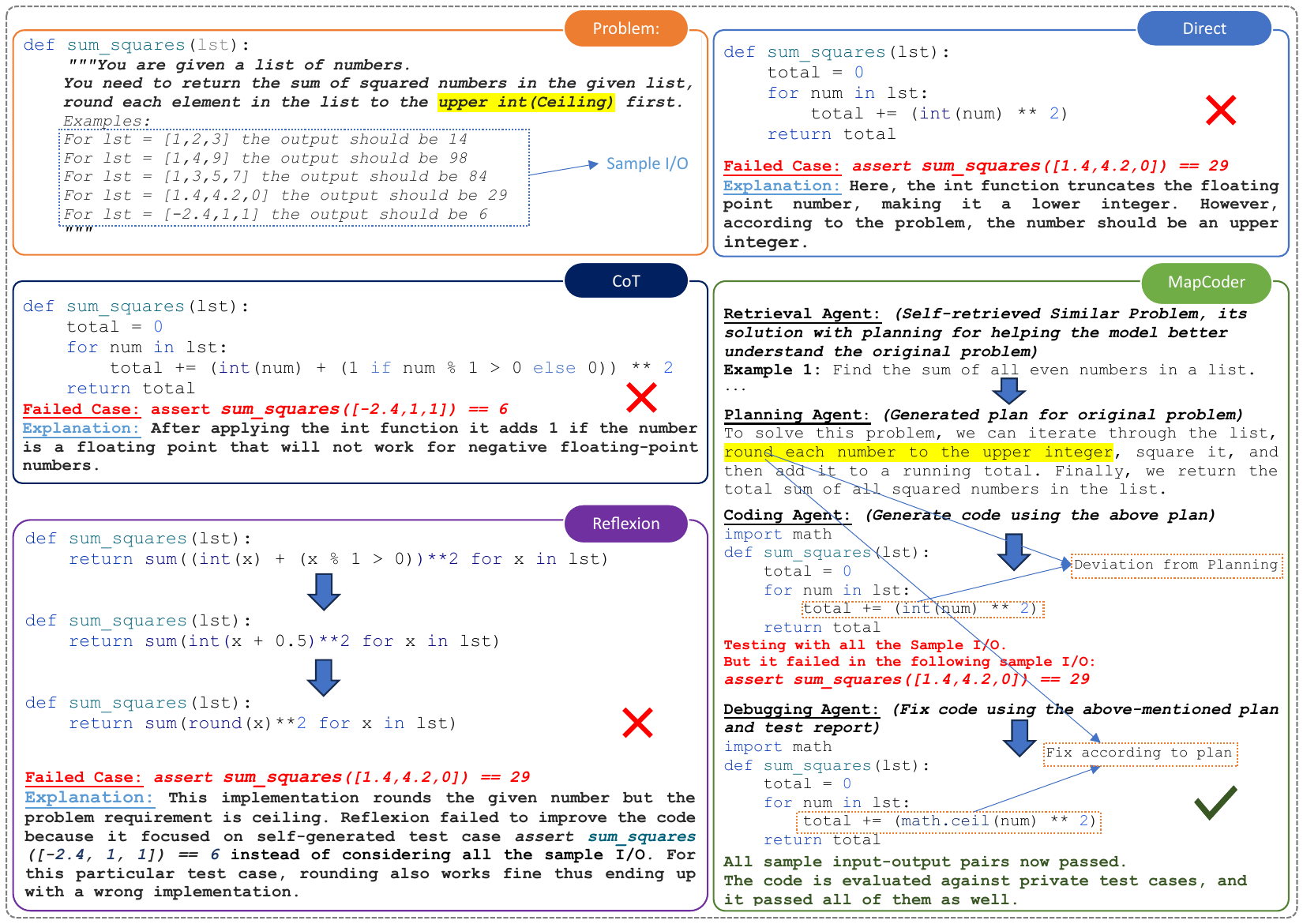}
    \vspace{-3mm}
    \caption{Example problem and solution generation using Direct, CoT, Reflexion, and \tool prompts. \tool explores high-utility plans first and uniquely features a plan-derived debugging for enhanced bug fixing.}
    \label{fig:qualitative-example}
    \vspace{-4mm}
\end{figure*}  

\smallskip
\subsection{Dynamic Agent Traversal}
\label{sec:agent-traverse}
The dynamic traversal in \tool begins with the \emph{Planning Agent}, which outputs the plans for the original problem with confidence scores. These plans are sorted, and the highest-scoring one is sent to the Coding Agent. The Coding Agent translates the plan into code, tested with sample I/Os. If all pass, the code is returned; otherwise, it's passed to \emph{Debugging Agent}. They attempt to rectify the code iteratively up to $t$ times. If successful, the code is returned; otherwise, responsibility shifts back to the \emph{Planning Agent} for the next highest confidence plan. This iterative process continues for $k$ iterations, reflecting a programmer's approach. We summarize our agent traversal in Algorithm \ref{alg:mapcoder} in Appendix. Our algorithm's complexity is $O(kt)$. An example illustrating \toolnospace's problem-solving compared to Direct, Chain-of-thought, and Reflexion approaches is in Figure \ref{fig:qualitative-example}.  All detailed prompts for each agent are in Appendix \ref{app:prompts}.

%% file: sections/experimental-setup.tex
\section{Experimental Setup}
\subsection{Datasets}
For extensive evaluation, we have used eight benchmark datasets: five from basic programming and three from complex competitive programming domains. Five basic programming datasets are: \textbf{HumanEval}~\cite{chen2021codex}, \textbf{HumanEval-ET}~\cite{dong2023codescore}, \textbf{EvalPlus}~\cite{evalplus}, \textbf{MBPP})~\cite{austin2021program}, and \textbf{MBPP-ET}~\cite{dong2023codescore}. HumanEval-ET, EvalPlus extend HumanEval and MBPP-ET comprehends MBPP by incorporating more test cases. The problem set size of HumanEval and MBPP (and their extensions) are 164 and 397, respectively. 
Due to the absence of sample I/O in MBPP and MBPP-ET, our approach for code moderation involves randomly removing one test-case from MBPP-ET for each problem and provide this test-case as a sample I/O for the problem. Importantly, this removed test-case is carefully selected to ensure mutual exclusivity from the hidden test sets in MBPP and MBPP-ET. Three competitive programming datasets are: Automated Programming Progress Standard (\textbf{APPS}), \textbf{xCodeEval} \cite{khan2023xcodeeval}, and \textbf{CodeContest}, where we have used 150, 106, and 156 problems, respectively, in our experiments.

\subsection{Baselines}
We have compared \tool with several baselines and state-of-the-art approaches. \textbf{Direct} Prompting instructs language models to generate code without explicit guidance, relying on their inherent capabilities of LLM. Chain of Thought Prompting (\textbf{CoT}) \cite{wei2022chain} breaks down problems into step-by-step solutions, enabling effective tackling of complex tasks. \textbf{Self-Planning} Prompting \cite{jiang2023self} divides the code generation task into planning and implementation phases. \textbf{Analogical Reasoning} Prompting \cite{yasunaga2023large} instructs models to recall relevant problems from training data. \textbf{Reflexion} \cite{shinn2023reflexion} provides verbal feedback to enhance solutions based on unit test results. \textbf{Self-collaboration} \cite{dong2023selfcollaboration} proposes a framework where different LLMs act as analyst, coder, and tester to cooperatively generate code for complex tasks, achieving better performance than directly using a single LLM. \textbf{AlphaCodium} \cite{ridnik2024code} iteratively refines code based on AI-generated input-output tests.

\input{tables/six-dataset-result}

\subsection{Foundation Models, Evaluation Metric, $k$, and $t$}
With  $k = t = 5$ in HumanEval, and $k = t = 3$ for others, we evaluate all the datasets using \href{https://platform.openai.com/docs/models/gpt-3-5-turbo}{ChatGPT (gpt-3.5-turbo-1106)}, \href{https://platform.openai.com/docs/models/gpt-4-and-gpt-4-turbo}{GPT-4 (gpt-4-1106-preview)} from OpenAI and \href{https://deepmind.google/technologies/gemini/#gemini-1.0}{Gemini Pro} from Google. We have also evaluated our method using an open-source LLM, Mistral-7B-instruct. We have used the Pass@k evaluation metric, where the model is considered successful if at least one of the $k$ generated solutions is correct.

%% file: tables/six-dataset-result.tex
\begin{table*}[h]
    \centering
    \begin{tabular}{c}
    \hspace{-2.5mm}
    \includegraphics[width=0.99\textwidth]{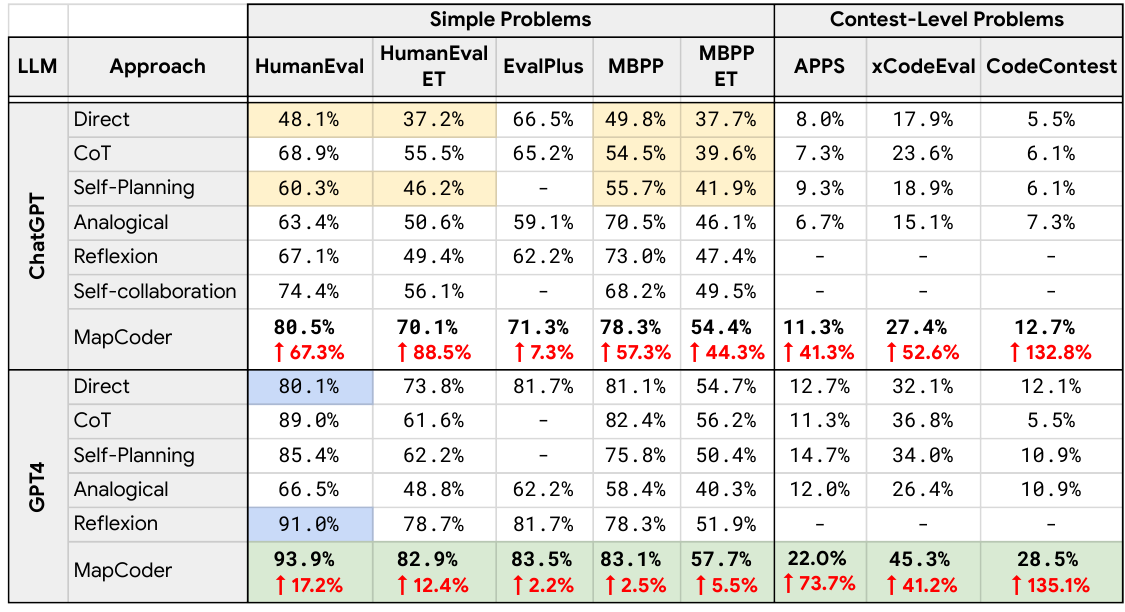}
    \end{tabular}
    \vspace{-3mm}
    \caption{Pass@1 results for different approaches. The results of the yellow and blue colored cells are obtained from \citet{jiang2023self} and \citet{shinn2023reflexion}, respectively. The results of the Self-collaboration \citet{dong2023selfcollaboration} paper are collected from their paper. The green texts indicate the state-of-the-art results, and the red text is gain over Direct Prompting approach.}
    \label{tab:six-dataset-result}
\end{table*}

%% file: sections/results.tex
\section{Results}
\label{sec:results}
In this section, we evaluate the code generation capabilities of our framework, \toolnospace, for competitive problem solving. Our experimental results are reported in Table \ref{tab:six-dataset-result}. 
Overall, \tool shows a tremendous excellence in code generation, significantly outperforms all baselines, and achieves new state-of-the-art results in all benchmarks. In general the scales with GPT-4 are higher than ChatGPT.

\subsection{Performance on basic code generation}
The highest scale of performance (Pass@1) scores are observed in simple program synthesis tasks like HumanEval, MBPP in Table \ref{tab:six-dataset-result}. 
Though with the simpler problem (non-contests) datasets such as HumanEval, HumanEval-ET, the current state-of-the-art method, Reflexion \cite{shinn2023reflexion} perform reasonably well, this approach does not generalize across varying datasets depicting a wide variety of problems. 
Self-reflection techniques enhance GPT-4's performance on HumanEval  but result in a 3\% decrease on the MBPP dataset. Similarly, with ChatGPT, there's a notable 26.3\% drop in performance where in several cases their AI generated test cases are incorrect. We observe that 8\% of failures in HumanEval and 15\% in MBPP is caused by  their AI generates incorrect test cases while our approach is independent of AI test cases, and consistently improves code generations in general. Consequently, even in HumanEval, with GPT-4, our Pass@1 surpasses  Reflexion by $\sim$3\%. On top, in all four simple programming datasets, \tool enhances the Direct prompting significantly with a maximum of 88\% on HumanEvalET by ChatGPT. 

\begin{figure*}[h]
    \centering
    \includegraphics[width=0.95\textwidth]{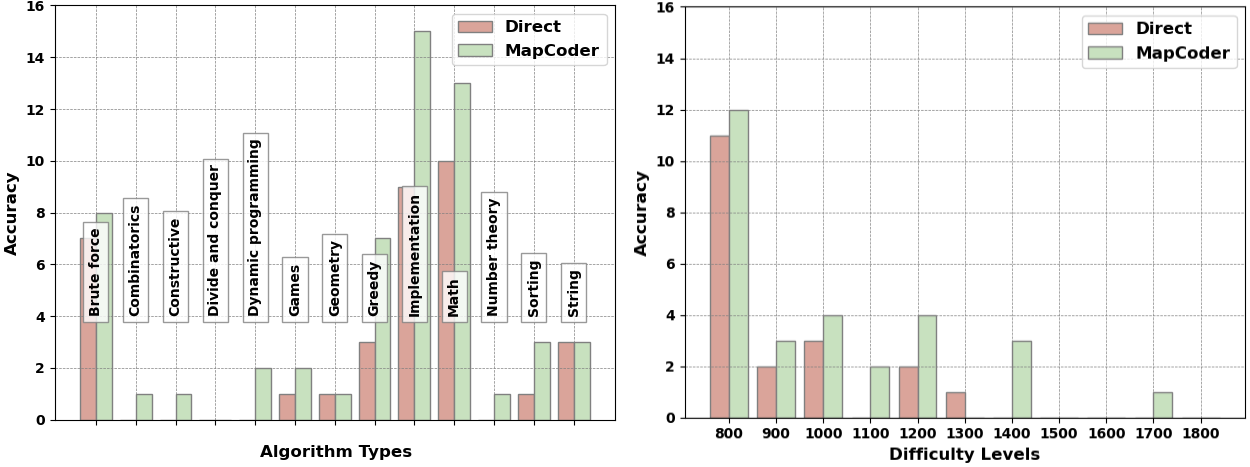}
    \vspace{-2mm}
    \caption{The number of correct answers wrt algorithm types (tags) and difficulty levels (xCodeEval dataset).} 
    \label{fig:xcode-aldo-diff}
    \vspace{-4mm}
\end{figure*}

\subsection{Performance on competitive problem solving}
The significance of \tool shines through clearly when evaluated in competitive problem-solving contexts. Across datasets such as APPS, xCodeEval, and CodeContests, \tool demonstrates substantial enhancements over Direct prompting methods, with improvements of 41.3\%, 52.6\%, and 132.8\% for ChatGPT, and 73.7\%, 41.2\%, and 135.1\% for GPT4, respectively. Notably, the most challenging datasets are APPS and CodeContest, where \toolnospace's performance stands out prominently. We deliberately compare against strong baselines on these datasets, regardless of whether they are prompt-based or not. 
Importantly, on CodeContest our Pass@1 results match the Pass@5 scores of the concurrent state-of-the-art model AlphaCodium \citep{ridnik2024code}: 28.5\% vs. their 29\% (see Table \ref{tab:cc-dataset-results}). Furthermore, our Pass@5 results demonstrate an additional improvement of 12.8\%. On APPS, \tool consistently surpasses the Pass@1 scores of all baseline prompts for both ChatGPT and GPT-4.

\input{tables/cc-results}

\subsection{Performance with Varying Difficulty Levels}
\begin{figure}[h]
    \centering
    \vspace{-3mm}
    \hspace*{-0.3cm}
    \includegraphics[width=.49\textwidth]{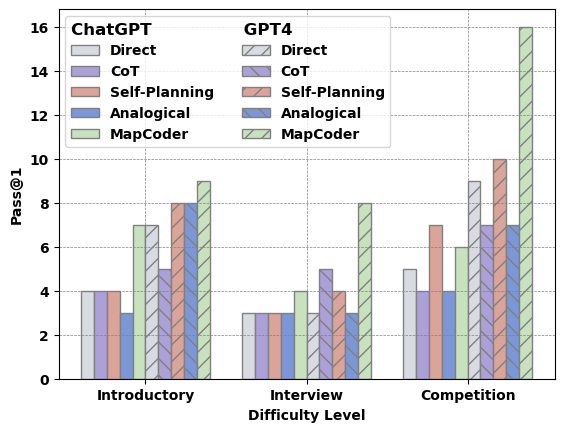}
    \vspace{-8mm}
    \caption{Performance vs problem types (APPS). }
    \label{fig:apps-difficulty-wise-plot}
    \vspace{-3mm}
\end{figure}

The APPS dataset comprises problems categorized into three difficulty levels: (i) Introductory, (ii) Interview, and (iii) Competition. Figure \ref{fig:apps-difficulty-wise-plot} illustrates the performance of various competitive approaches for these three categories. The results reveal that our \tool excels across all problem categories, with highest gain in competitive problem-solving indicating its superior  code generation capabilities in general, and on top, remarkable effectiveness in competitive problem-solving.  In order to gather more understanding on what algorithm problems it's capable of solving and in fact much difficulty level it can solve,  we have also conducted a comparison between \tool and the Direct approach, considering the difficulty levels\footnote{Difficulty levels in xCodeEval dataset represents an integer number, a higher value means more difficult problem} and tags\footnote{Tags in xCodeEval dataset represents algorithm type that can be used to solve the problem i.e., greedy, dp, brute-force, constructive, and so on.} present in the xCodeEval dataset. The results of this comparison are depicted in Figure \ref{fig:xcode-aldo-diff}. This comparison showcases that \tool is effective across various algorithm types and exhibits superior performance even in higher difficulty levels, compared to the Direct approach. However, beyond (mid-level: difficulties>1000), its gains are still limited. 

\subsection{Performance Across Different LLMs}
To show the robustness of \tool across various LLMs, we evaluate \tool using Gemini Pro, a different family of SoTA LLM in Table~\ref{tab:gemini-results}. We also evaluate \tool using an open-source LLM Mistral-7B instruct in Table~\ref{tab:mistral-results}. As expected, our method shows performance gains over other baseline approaches in equitable trends on both simple (HumanEval) and contest-level problems (CodeContest). 

\input{tables/gemini-results}
\input{tables/mistral-results}

\subsection{Performance Across Different Programming Languages}
\begin{figure}[h]
    \centering
    \includegraphics[width=0.4\textwidth]{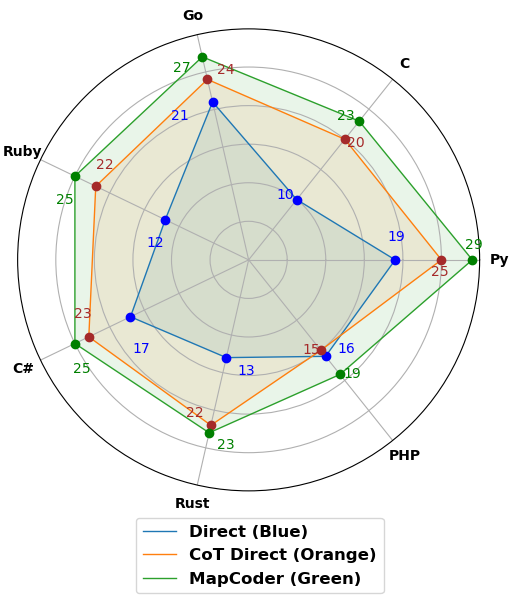}
    \vspace{-3mm}
    \caption{The number of correct answers wrt different programming languages (xCodeEval dataset). }
    \label{fig:xcode-multi-lingual}
    \vspace{-4mm}
\end{figure}

Furthermore, we evaluate model performances using \tool across different programming languages. We utilize the xCodeEval dataset, which features multiple languages. Figure \ref{fig:xcode-multi-lingual} shows that consistent proficiency across different programming languages is achieved by \tool with respect to baselines.

%% file: tables/cc-results.tex
\begin{table}[h]
    \centering
    \begin{tabular}{c}
    \hspace*{-0.3cm}
    \includegraphics[width=0.48\textwidth]{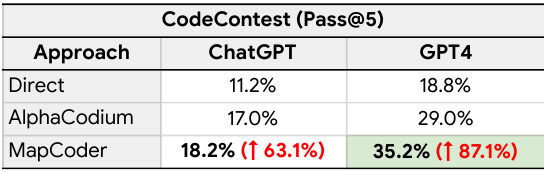}
    \end{tabular}
    \caption{Pass@5 results on CodeContest dataset. AlphCodium result are from \citet{ridnik2024code}. The green cells indicate the SoTA and the red text indicates  improvement w.r.t Direct approach.}
    \label{tab:cc-dataset-results}
\end{table}

%% file: tables/gemini-results.tex
\begin{table}[h]
    \centering
    \begin{tabular}{c}
    \hspace*{-0.35cm}
    \includegraphics[width=0.45\textwidth]{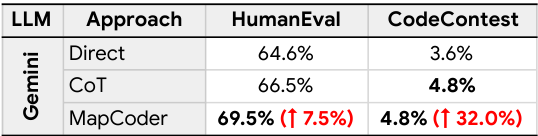}
    \end{tabular}
    \vspace{-3mm}
    \caption{Pass@1 results with using Gemini Pro. The red text is gain over Direct Prompting approach. }
    \label{tab:gemini-results}
\end{table}

%% file: tables/mistral-results.tex
\begin{table}[h]
    \centering
    \vspace{-1mm}
    \begin{tabular}{c}
    \hspace*{-0.35cm}
    \includegraphics[width=0.45\textwidth]{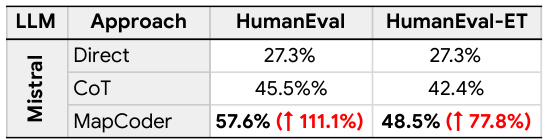}
    \end{tabular}
    \vspace{-3mm}
    \caption{Pass@1 results with using Mistral-7B-instruct. The red text is gain over Direct Prompting approach. }
    \label{tab:mistral-results}
    \vspace{-2mm}
\end{table}

%% file: sections/ablation-study.tex
\section{Ablations Studies and Analyses}
\label{sec:ablation-study}
We  present the ablation study of the \tool on HumanEval dataset as the problems are simpler and easy to diagnose by us humans. 

\subsection{Impact of Different Agents}
\label{sec:agent-impact-ablation}
We have also conducted a study by excluding certain agents from our \toolnospace, which helps us investigate each agent's impact in our whole pipeline. As expected, the results (Table \ref{tab:agent-disabling}) show that every agent has its role in the pipeline as turning off any agent decreases the performance of \toolnospace. Furthermore, we observe that the Debugging Agent has the most significant impact on the pipeline, as evidenced by a performance drop of 17.5\% when excluding this agent exclusively, and an avg performance drop of 24.83\% in all cases. The \emph{Planning agent} has the second best important with avg drop of 16.7\% in all cases. 
In Table \ref{tab:agent-disabling}), we perform an ablation study of our multi-agent framework investigate each agent's impact in our whole pipeline.

\input{tables/ablationwa}
\subsection{Qualitative Example}
To verify the above numerical significance, and to understand how our method enhance the code generation, we have performed a qualitative analysis to find the underlying reason for the superior performance of \tool over other competitive prompting approaches. An example problem and the output with the explanation of Direct, CoT, Reflexion, and \tool prompting is shown in Figure \ref{fig:qualitative-example}. This example demonstrates how the \emph{Debugging Agent} fixes the bugs leveraging the plan as a guide from the \emph{Planning Agent}. This verifies the impact of these two most significant agents. We present more detailed examples in Appendix.

\subsection{Impact of $k$ and $t$}
\tool involves two hyper-parameters: the number of self-retrieved exemplars, $k$, and the number of debugging attempts, $t$. Our findings (Table \ref{tab:ablation-k-t-results}) reveal that higher $k$, $t$  is proportionate performance gain at the expense of time.  
\input{tables/ablationkt}

\input{tables/ablation-att}

\subsection{Impact of Number of Sample I/Os}
Given the limited number of sample I/Os in the HumanEval dataset (average of 2.82 per problem), we supplemented it with an additional 5 sample I/Os from the HumanEval-ET dataset. Experiments with this augmented set showed an 1.5\% performance gain. \\


\subsection{Error Analysis and Challenges}
Although \tool demonstrates strong performance compared to other methods, it faces challenges in certain algorithmic domains. For example, Figure \ref{fig:xcode-aldo-diff} illustrates \toolnospace's reduced performance on more difficult problems requiring precise problem understanding and concrete planning—capabilities still lacking in LLMs.
In the xCodeEval dataset (see Figure \ref{fig:xcode-aldo-diff}), it solves a limited number of problems in categories like Combinatorics, Constructive, Number Theory, Divide and Conquer, and Dynamic Programming (DP). Manual inspection of five DP category problems reveals occasional misinterpretation of problems, attempts to solve using greedy or brute-force approaches, and struggles with accurate DP table construction when recognizing the need for a DP solution. 

%% file: tables/ablationwa.tex
\begin{table}[h]
    \centering
    \hspace{-6mm}
    \resizebox{0.52\textwidth}{!}{
    \begin{tabular}{c}
    \hspace*{-0.3cm}
    \includegraphics[width=0.48\textwidth]{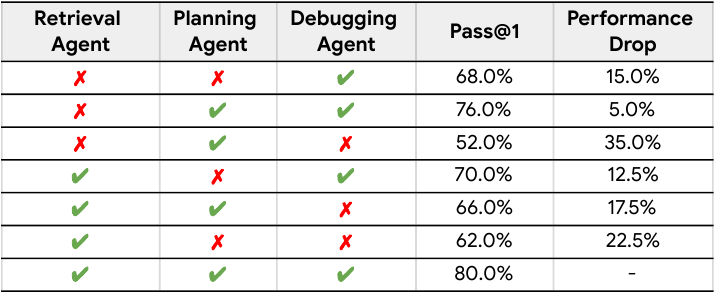}
    \end{tabular}
    }
    \vspace{-2mm}
    \caption{Pass@1 results for different versions of \tool (by using ChatGPT on HumanEval dataset).}
    \label{tab:agent-disabling}
    \vspace{-2mm}
\end{table}

%% file: tables/ablationkt.tex
\begin{table}[h]
    \centering
    \vspace{-2mm}
    \resizebox{0.52\textwidth}{!}{
    \begin{tabular}{c}
    \hspace*{-0.3cm}
    \includegraphics[width=0.48\textwidth]{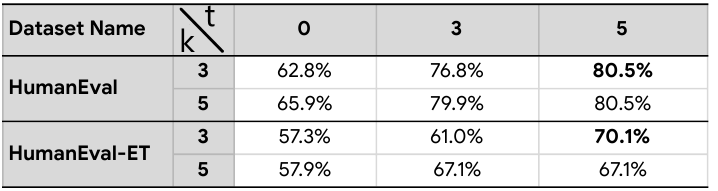}
    \end{tabular}
    }
    \vspace{-2mm}
    \caption{Pass@1 results by varying $k$ and $t$.}
    \label{tab:ablation-k-t-results}
    \vspace{-2mm}
\end{table}

%% file: tables/ablation-att.tex
\begin{table*}[t]
    \centering
    \vspace{-5mm}
    \begin{tabular}{c}
    \hspace*{-0.3cm}
    \includegraphics[width=0.80\textwidth]{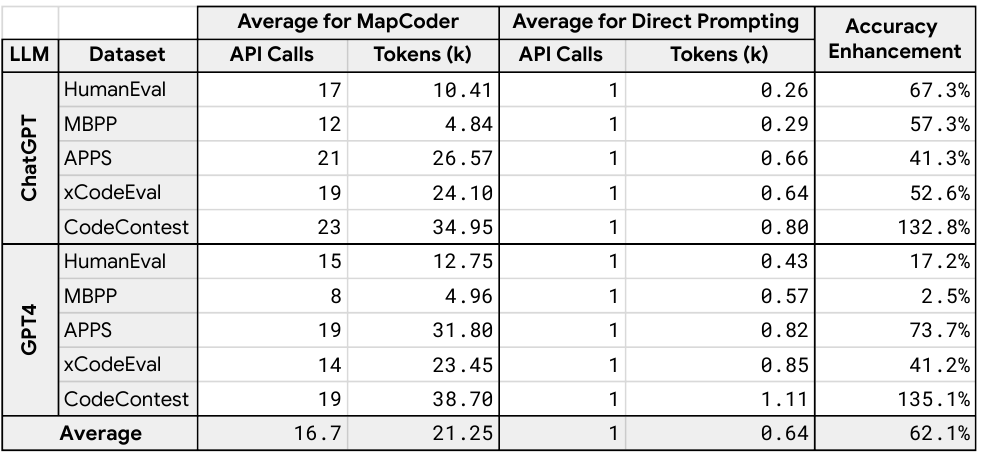}
    \end{tabular}
    \vspace{-3mm}
    \caption{Average number of API calls, thousands of tokens used, required time in minutes to get the API response. }
    \label{tab:ablation-att}
    \vspace{-4mm}
\end{table*}

%% file: sections/conclusion.tex
\section{Conclusion and Future Work}
In this paper, we introduce \toolnospace, a novel framework for effective code generation in complex problem-solving tasks, leveraging the multi-agent prompting capabilities of LLMs. \tool captures the complete problem-solving cycle by employing four agents - retrieval, planning, coding, and debugging - which dynamically interact to produce high-quality outputs. Evaluation across major benchmarks, including basic and competitive programming datasets, demonstrates \toolnospace's consistent outperformance of well-established baselines and SoTA approaches across various metrics. Future work aims to extend this approach to other domains like question answering and mathematical reasoning, expanding its scope and impact.

%% file: sections/limitations.tex
\section{Limitations}
Among the limitations of our work, firstly, \tool generates a large number of tokens, which may pose challenges in resource-constrained environments. Table \ref{tab:ablation-att} shows the number of average API calls and token consumption with the default $k$ and $t$ (i.e., with respect to the reported performance)  while  Table \ref{tab:ablation-k-t-results}) shows how $k$, $t$ can be adjusted to proportionate the performance gain at the expense of time/token. We have not addressed the problem of minimizing tokens/API-calls in this paper and leave it for future works. Secondly, our method currently relies on sample input-output (I/O) pairs for bug fixing. Although sample I/Os provide valuable insights for LLMs' code generation, their limited number may not always capture the full spectrum of possible test cases. Consequently, enhancing the quality of additional test case generation could reduce our reliance on sample I/Os and further improve the robustness of our approach. Additionally, future exploration of open-source code generation models, such as CodeLLaMa, 
LLaMa3, Mixtral 8x7B could offer valuable insights and potential enhancements to our approach. Another important concern is that while running machine-generated code, it is advisable to run it inside a sandbox to avoid any potential risks.

%% file: sections/ack.tex

%% file: appendix/appendix.tex
\vspace{8mm}
\appendix

\noindent {\LARGE \textbf{Appendix}} \\

\section{Algorithm of \ourapproach}
\label{alg:mapcoder}

Algorithm~1 shows the pseudo-code of our prompting technique.

\begin{algorithm}[h]
\small
\caption{\tool}
\begin{algorithmic}[1]
\State $k \gets$ number of self-retrieved exemplars
\State $t \gets$ number of debugging attempts \\

\State $exemplars \gets$ RetrivalAgent($k$) \\ 

\State $plans \gets$ empty array of size $k$
\For{$example$ in $exemplars$} 
    \State $plans[i] \gets$ PlanningAgent($example$)
\EndFor \\

\State $plans \gets$ SortByConfidence($plans$) \\
\For{$i \gets 1$ to $k$}
    \State $code \gets$ CodingAgent($code$, $plan[i]$) 
    \State $passed,\space log \gets$ test($code$, $sample\_io$) 
    \If{$passed$}
        \State Return $code$
    \Else
        \For{$j \gets 1$ to $t$}
            \State $code \gets$ DebuggingAgent($code$, $log$) 
            \State $passed,\space log \gets$ test($code$, $sample\_io$)
            \If{$passed$}
                \State Return $code$
            \EndIf
        \EndFor
    \EndIf
\EndFor
\State Return $code$
\end{algorithmic}
\end{algorithm}

\section{Details Promptings of \ourapproach}
\label{app:prompts}
\begin{figure*}[h]
    \centering
    \includegraphics[width=0.99\textwidth]{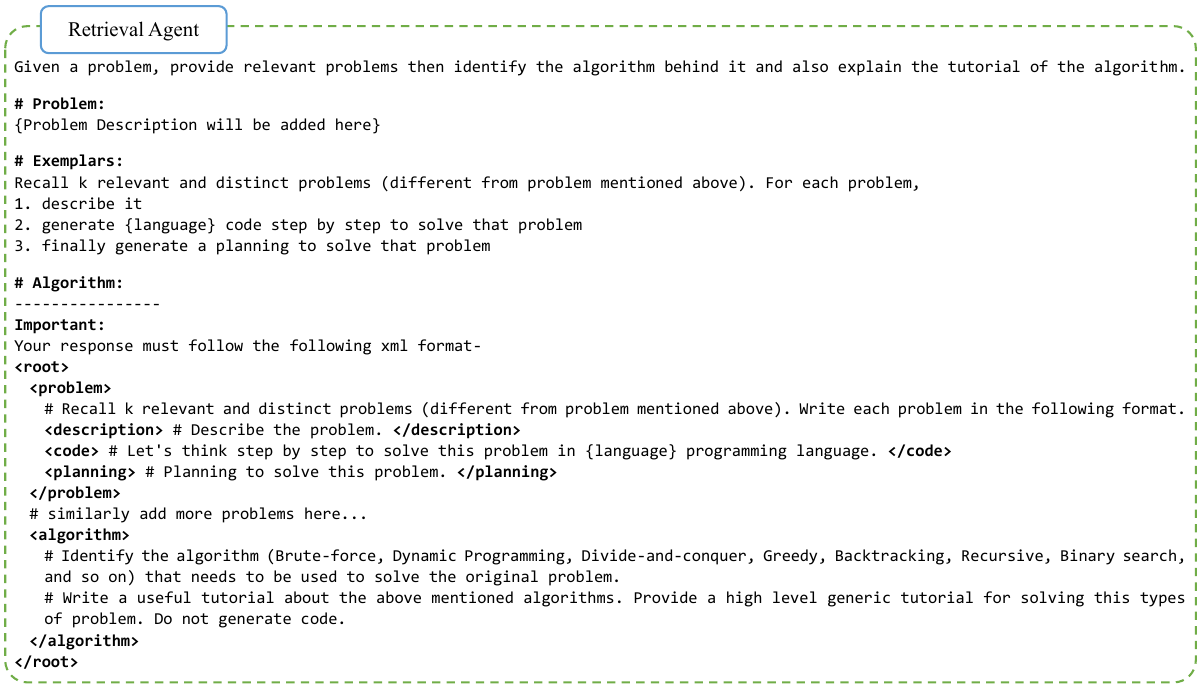}
    \caption{Prompt for self-retrieval Agent. }
    \label{fig:prompt-agent-1}
\end{figure*}

\begin{figure*}[h]
    \centering
    \includegraphics[width=0.99\textwidth]{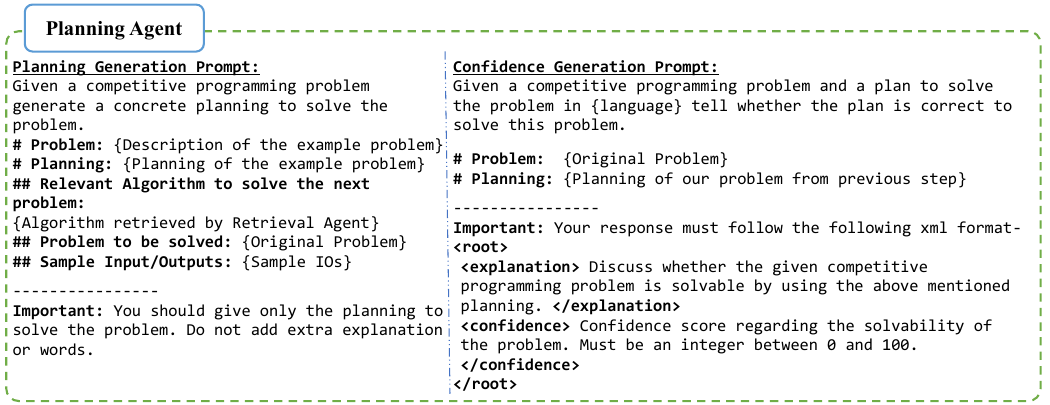}
    \caption{Prompt for Planning Agent. The example problems that are mentioned in this figure will come from the Retrieval Agent.}
    \label{fig:prompt-agent-2}
\end{figure*}

\begin{figure*}[h]
    \centering
    \includegraphics[width=0.99\textwidth]{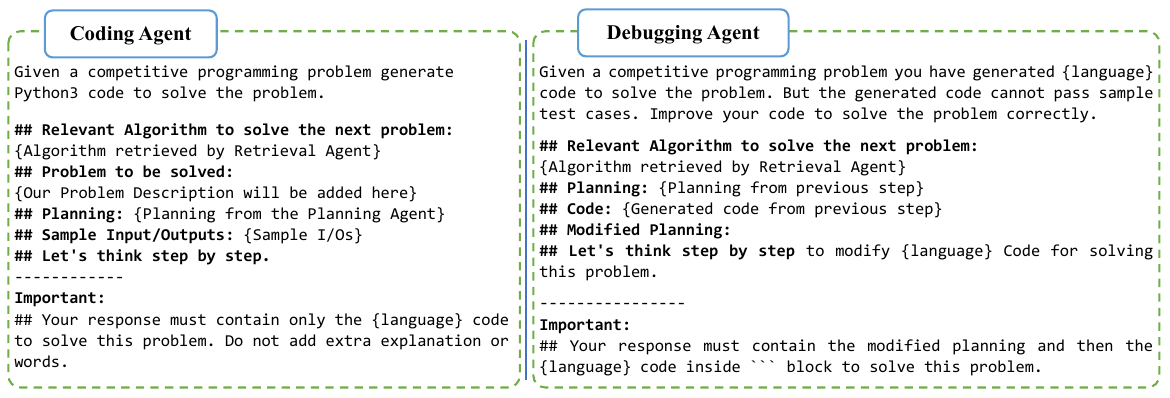}
    \caption{
    Prompt for Coding and Debugging Agent.}
    \label{fig:prompt-agent-3-4}
\end{figure*}

The detailed prompting of the Retrieval Agent, Planning Agent, Coding Agent, and Debugging Agent are shown in Figure \ref{fig:prompt-agent-1}, \ref{fig:prompt-agent-2}, and \ref{fig:prompt-agent-3-4} respectively. Note that we adopt a specific sequence of instructions in the prompt for Retrieval Agent which is a crucial design choice. 

\section{Example Problem}
\label{app:example-problem}
Two complete examples of how \tool works by showing all the prompts and responses for all four agents is given in this \href{https://github.com/Md-Ashraful-Pramanik/mapcoder.github.io/blob/master/files/example-prompt.pdf}{link}.

\newpage

